\documentclass[10pt,twocolumn,letterpaper]{article}

\usepackage[pagenumbers]{cvpr} 
\usepackage{multirow}

\newcommand{\bp}{{\bf p}}

\definecolor{cvprblue}{rgb}{0.21,0.49,0.74}
\usepackage[pagebackref,breaklinks,colorlinks,allcolors=cvprblue]{hyperref}

\def\paperID{7179} 
\def\confName{CVPR}
\def\confYear{2025}

\title{Compression of Large-Scale 3D Point Clouds Based on Joint Optimization of Point Sampling and Feature Extraction}

\author{
Jae-Young Yim \quad Jae-Young Sim \\
\\
Ulsan National Institute of Science and Technology~(UNIST) \\
{\tt\small \{yimjae0, jysim\}@unist.ac.kr}
}

\begin{document}
\maketitle

\begin{abstract}
Large-scale 3D point clouds (LS3DPC) obtained by LiDAR scanners require huge storage space and transmission bandwidth due to a large amount of data. The existing methods of LS3DPC compression separately perform rule-based point sampling and learnable feature extraction, and hence achieve limited compression performance. In this paper, we propose a fully end-to-end training framework for LS3DPC compression where the point sampling and the feature extraction are jointly optimized in terms of the rate and distortion losses. To this end, we first make the point sampling module to be trainable such that an optimal position of the downsampled point is estimated via aggregation with learnable weights. We also develop a reliable point reconstruction scheme that adaptively aggregates the expanded candidate points to refine the positions of upsampled points. Experimental results evaluated on the SemanticKITTI and nuScenes datasets show that the proposed method achieves significantly higher compression ratios compared with the existing state-of-the-art methods.
\end{abstract}

\section{Introduction}
\label{sec:intro}

LiDAR (Light Detection and Ranging) scanners have been used to generate large-scale 3D point clouds (LS3DPC) capturing real-world scenes, where a single scene is usually composed of millions or tens of millions of points. The principle of LiDAR is time-of-flight (TOF) imaging. The scanner first emits laser pulses associated with certain azimuthal and polar resolutions while rotating around the vertical axis. Then the scanner receives the echo pulses reflected on the surface of real-world objects. The distance from the location of scanner to the object is estimated by calculating the traveling time of laser pulses. The obtained LS3DPC are dense and accurate and have been used in various applications such as terrain modelling~\cite{terrain}, object detection~\cite{pointpillars}, and navigation~\cite{navigation}.

A vast volume of LS3DPC data requires efficient compression techniques to alleviate the storage space and transmission bandwidth.~Lossy compression techniques have been actively developed.~Whereas the traditional methods~\cite{regionpcgc, dyadic, pointvote} provide relatively lower compression performance,~the learning-based methods~\cite{pointnetbased, pointnet2based, depoco, dpcc, 3qnet, mslpcc}, especially those using deep learning techniques, achieve superior performance over the traditional ones.~The pioneering architectures of PointNet~\cite{pointnet} and PointNet++~\cite{pointnet2} have been used to construct autoencoders specifically designed for point cloud data compression~\cite{pointnetbased, pointnet2based}. DEPOCO~\cite{depoco} utilizes KPConv~\cite{kpconv} to effectively capture the local structures and features of 3D point clouds. D-PCC~\cite{dpcc}, the state-of-the-art method, enhances the diversity of features and mitigates the issue of locally concentrated point distribution. Scalable compression frameworks such as 3QNet~\cite{3qnet} and msLPCC~\cite{mslpcc} have been also proposed. 

\begin{figure}[t]
  \centering
  \includegraphics[width=1\linewidth]{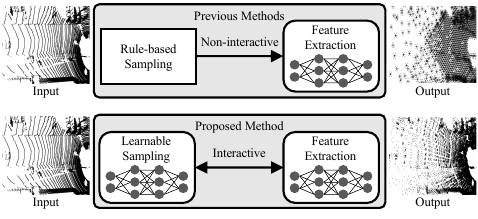}
  \caption{The concept of the proposed joint optimization framework between learnable point sampling and feature extraction, compared to the existing methods where the rule-based sampling is performed separately from the feature extraction.}
  \label{fig:fig1}
\end{figure}
However, as shown in Figure~\ref{fig:fig1}, most of the existing methods simply employ rule-based point downsampling followed by learning based feature extraction, where these sequential operations are conducted separately. It means that the downsampling is not influenced by the feature extraction and vice versa. In such a case, the spatial characteristics across the points, specifically beneficial to compression, may be often overlooked during the rule-based downsampling that selects points without considering the compression performance. Furthermore, the extracted features aggregated onto the downsampled points may not optimal to improve the compression performance as well.

To overcome the aforementioned limitations of the existing methods, we propose a fully end-to-end training framework which jointly optimizes the point downsampling and feature extraction, as compared in Figure~\ref{fig:fig1}.~Specifically, we uniformly partition an input LS3DPC model based on the range image domain to effectively handle the irregular densities of point distribution. We devise a learnable point sampling network that aggregates the points in each partition into a single point via learnable weights. Moreover, the deep network for point sampling is jointly trained with that of the feature extraction in order to select an optimal location of the downsampled point while extracting an optimal feature in terms of the compression performance. At the decoder side, we also devise an expansion and fusion based refinement module for reliable point reconstruction, which first expands candidate points and then estimates an optimal reconstruction point based on an ensemble technique. Experimental results demonstrate that the proposed method dramatically improves the performance of LS3DPC compression compared with the existing state-of-the-art methods.

The main contributions of this paper are summarized as follows.
\begin{itemize}
  \item To the best of our knowledge, we first propose a fully end-to-end training framework for the compression of LS3DPC, where a learnable point sampling network is jointly optimized with a feature extraction network in terms of the compression performance.
  \item We devise an expansion and fusion scheme for point refinement to further improve the quality of the reconstructed point clouds. 
  \item Experimental results demonstrate that the proposed method significantly improves the rate-distortion performance of the existing state-of-the-art methods.
\end{itemize}

\section{Related Work}
\label{sec:related}

\subsection{Feature Extraction for 3D Point Clouds}
LS3DPC exhibit unstructured and irregular distributions which make it challenging to extract geometric features. PointNet~\cite{pointnet}, a pioneering architecture, uses symmetric functions like global max pooling to ensure the permutation invariance for disordered points. PointNet++~\cite{pointnet2}, an extended version of PointNet, performs sampling and grouping steps to handle varying point densities and hierarchically processes the point clouds to better capture the local structures. Research is also underway to develop CNN-like operations on 3D point clouds. DGCNN~\cite{dgcnn} introduces dynamic graph convolution which updates the graph in feature space to capture semantically similar geometric structures. KPConv~\cite{kpconv} is a 3D convolutional neural network which uses rigid or deformable kernels of points to learn various 3D patterns. Attention mechanisms~\cite{transformer, vit} have been widely used to adaptively assign higher weights to more important regions or elements when processing images and texts. Similarly, Point Transformer~\cite{pointtransformer} leverages the attention mechanisms to understand the relationships between the points far way from each other without relying on traditional convolutional structures.

\begin{figure*}[t]
  \centering
	\includegraphics[scale=0.5]{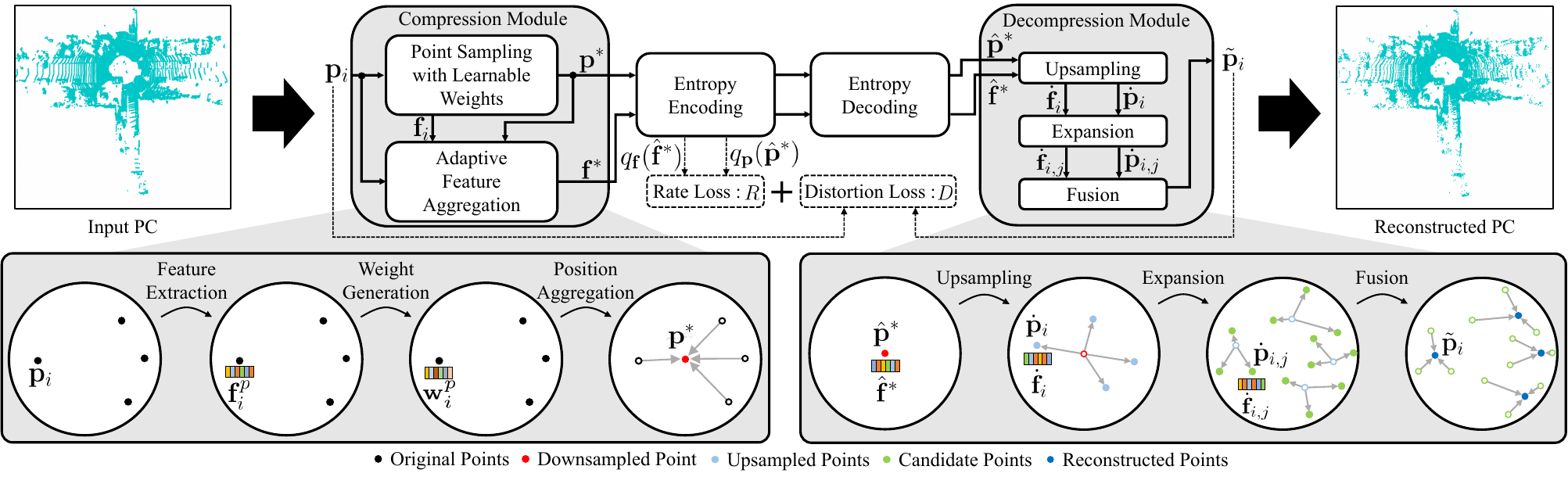}
  \caption{Overall procedures of the proposed method.}
  \label{fig:fig2}
\end{figure*}

\subsection{Lossless Compression for 3D Point Clouds}
Lossless compression techniques for 3D point clouds first quantize the position of points which are then encoded without distortion. Traditional methods~\cite{rangetra1, rangetra2} first project an input LS3DPC model onto the range image domain where the image compression techniques are applied. Deep learning techniques~\cite{riddle, rangeentropy} have been also adopted to improve the compression performance.~Tree structures are often used to represent 3D point cloud data like the MPEG standard G-PCC~\cite{gpcc}.~Whereas handcrafted entropy models are used~\cite{traditional1, traditional2, gpcc, draco, layerwise}, more recent studies~\cite{octsqueeze, voxelcontext, muscle, sparsepcgc, octformer, efficient, scp, isolated, multiscale} use learning-based entropy models to more accurately predict the data distribution. 

\subsection{Lossy Compression for 3D Point Clouds}
With the advent of learned image compression techniques~\cite{end, hyperprior, joint} alongside the feature extractors~\cite{pointnet, pointnet2, kpconv} for 3D point clouds, lossy compression for 3D point clouds has been attracting much attention~\cite{pointnetbased, pointnet2based, pcgcv2, pccgeocnnv2, dpcc, depoco, 3qnet, mslpcc} which are categorized into voxel-based approaches~\cite{pcgcv1, pcgcv2, pccgeocnnv1, pccgeocnnv2, sparsepcgc} and point-based approaches~\cite{pointnetbased, pointnet2based, depoco, dpcc, 3qnet, mslpcc} depending on the input data representation. 

\subsubsection{Voxel-based Approaches} Voxel-based methods first convert an input point cloud model into the voxels to create a structured representation for 3D convolution operations. The methods~\cite{pccgeocnnv1, pccgeocnnv2, pcgcv1} using the conventional voxelization technique suffer from high memory consumption and computational complexity. To address this issue, advanced methods~\cite{pcgcv2, sparsepcgc} convert voxels into 3D sparse tensors using 3D sparse convolution reducing the memory usage. However, these methods still face the challenges of precision loss especially in LS3DPC captured by LiDAR scanners.

\subsubsection{Point-based Approaches} On the other hand, the initially captured raw point clouds are directly used. The autoencoders for point clouds compression have been developed by using the PointNet and PointNet++ architectures in~\cite{pointnetbased} and~\cite{pointnet2based}, respectively. DEPOCO~\cite{depoco} employs KPConv to process the dense point clouds by capturing spatially local structures and features. The state-of-the-art method, D-PCC~\cite{dpcc}, integrates a transformer~\cite{pointtransformer} into the compression process by aggregating multiple embeddings from the previous layer into the sampled points. Also, subpoint convolution operation alleviates the effect of locally clustered points and thus achieves superior performance of compression. Scalable compression also belongs to the lossy compression. Additionally, 3QNet~\cite{3qnet} permits the model to adapt to various compression rates in testing scenarios without retraining. msLPCC~\cite{mslpcc} decomposes an input 3D point cloud model into several downsampled parts to incrementally compress the data by incorporating the features from three modalities of depth, segmentation, and points.

\section{Proposed Method}


Most of the existing methods typically perform rule-based point downsampling and learning-based feature extraction separately during the encoding process. Therefore, they are limited to achieve optimal compression performance.~On the contrary, we devise a learning framework for point downsampling which is jointly optimized with the feature extraction in an end-to-end training manner. Figure~\ref{fig:fig2} presents the overall framework of the proposed method. We partition an input LS3DPC model into local patches based on the range image domain, where the points in each patch are downsampled into a single point. We determine an optimal location for the sampled point via learnable weights, and adaptively aggregate the deep features. Both procedures are jointly optimized to minimize the rate and distortion losses. The resulting downsampled point and the aggregated feature are entropy encoded. At the decompression phase, we initially upsample the points up to the same number of the original points, which are then further upsampled to expand the candidates. The expanded candidates are aggregated together to reliably refine the positions of the reconstructed points. 



\subsection{Point Sampling with Learnable Weights}

\begin{figure}[t]
	\centering
	\includegraphics[width=1\linewidth]{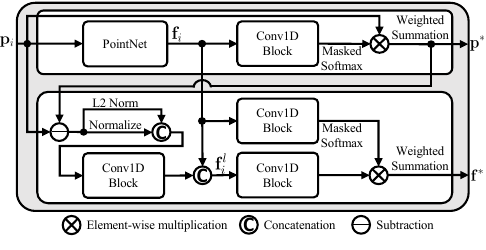}
	\caption{Network architecture of the compression module where the point sampling and feature aggregation are jointly trained.}
	\label{fig:fig3}
\end{figure}
We partition the LS3DPC along the azimuthal and polar directions reflecting the principle of LiDAR. 
Specifically, an input LS3DPC model is first projected onto the range image domain where each pixel corresponds to a 3D point associated with a certain pair of azimuthal and polar angles. The range image is then partitioned into equal-sized local patches where the set of points associated with each patch is used as an input to deep networks.~Note that we randomly select valid points within a same patch and assign them to the empty pixels which have no echo pulses. We employ a flag bit to classify such redundant points from the valid points to prevent the training bias. The patches with no valid points are excluded from the training.


The existing methods~\cite{pointnetbased, pointnet2based, depoco, dpcc, 3qnet, mslpcc} based on autoencoder architectures conduct rule-based point downsampling such as Farthest Point Sampling (FPS) or Random Sampling (RS). FPS iteratively selects the points that are farthest from the previously selected ones to ensure even distribution of downsampled points. RS randomly selects the points regardless of their spatial distribution.~Moreover, such downsampling is performed separately from the learning-based feature extraction during the encoding process.~Instead, we devise a learning-based framework for point downsampling which is jointly trained with the feature extraction network. Figure~\ref{fig:fig3} shows the network architecture for joint training between the point sampling and feature aggregation. Let ${\boldsymbol{\mathcal P}} = \left\{ \bp_i \right\}_{i=1}^N$ represent an input patch of $N$ points, which is supposed to be downsampled into a single point $\bp^*$. The location of $\bp^*$ is determined with the learnable weights ${\bf w}^p_i$'s given by
\begin{equation} \label{eq:eq2}
	\bp^{*} = \sum^{N}_{i=1} {{\bf w}^p_i} \otimes \bp_{i},
\end{equation}
where ${\bf w}^p_i$ is a $3 \times 1$ vector and $\otimes$ denotes the element-wise multiplication.

The learnable weights are obtained by using the features of points.~Let ${\bf f}_i$ be the feature vector of $\bp_i$ extracted by using the PointNet~\cite{pointnet}. We obtain a $3 \times 1$ vector of ${\bf f}^p_i$ by passing ${\bf f}_i$ through a convolution block followed by the exponentiation.~Then ${\bf w}^p_i$ is given by the masked softmax operation as
\begin{equation} \label{eq:eq3}
        {\bf w}^p_i = \frac{b_i \cdot {\bf f}^p_i} {\sum^{N}_{j=1} b_j \cdot {\bf f}^p_j},
\end{equation}
where the flag bit $b_i$ is set as $b_i=1$ when $\bp_{i}$ is a valid point, and $b_i=0$ for a redundant point to ensure that only the valid points are considered in the aggregation. Note that the proposed method learns the weights to sample $\bp^{*}$ at an optimal location beyond ${\boldsymbol{\mathcal P}}$ in terms of the compression performance by adaptively considering different contribution of the points.

\subsection{Adaptive Feature Aggregation} 
We compute the feature ${\bf f}^{*}$ for the downsampled point $\bp^*$ via adaptive feature aggregation as well. Specifically, the offset ${\Delta{\bf p}_i} = {\bf p}_i - \bp^*$ between an original point and the downsampled point is concatenated with the input feature ${\bf f}_i$ to form the combined feature ${\bf f}^l_i$. Then we aggregate the combined features adaptively by using the learnable weights ${\bf w}^f_i$'s as
\begin{equation} \label{eq:eq4}
{\bf f}^{*} = \sum^{N}_{i=1}{{\bf w}^f_i} \otimes {\bf f}^l_{i}.
\end{equation}
${\bf w}^f_i$ is a $C \times 1$ vector where $C$ is the number of feature channels, and computed in a similar way as~\eqref{eq:eq3}. Note that we attempt to compute an optimal feature in terms of the compression performance by adaptively aggregating the comprehensive point features.

\subsection{Reconstruction via Expansion and Fusion}

\begin{figure}[t]
	\centering
	\includegraphics[width=1\linewidth]{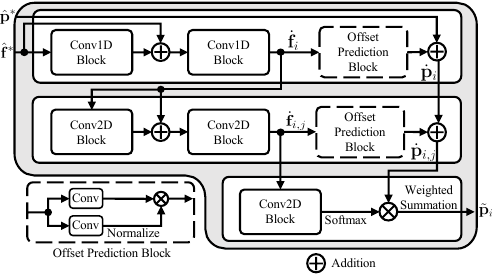}
	\caption{Network architecture of the decompression module where the point reconstruction is performed via expansion and fusion.}
	\label{fig:fig4}
\end{figure}
The optimal location ${\bf p}^{*}$ and the optimal feature ${\bf f}^{*}$ of the downsampled point are entropy encoded. As shown in Figure~\ref{fig:fig4}, the decoder reconstructs the original points $\tilde{\bf p}_i$'s from the entropy decoded point $\hat{\bp}^*$ and the feature $\hat{\bf f}^*$.
We initially upsample the feature vector $\hat{\bf f}^*$ to $N$ feature vectors, which are then used to predict the offsets between $\hat{\bp}^*$ and the original points. The restored offsets are added to $\hat{\bp}^*$ to obtain upsampled points $\dot{\bf p}_i$'s.

The initially upsampled points $\dot{\bf p}_i$'s undergo further refinement by applying an ensemble technique via expansion and fusion processes. In practice, we further apply the upsampling process to each $\dot{\bf p}_i$ again to generate $N_c$ candidate points $\dot{\bp}_{i,j}$'s, respectively, with the corresponding features $\dot{\bf f}_{i,j}$'s.
Finally, we estimate the reconstructed point $\tilde{\bf p}_i$ by adaptively aggregating the expanded candidate points with the learnable weights $\boldsymbol{\bf{w}}^{{p}}_{i,j}$'s given by
\begin{equation} \label{eq:eq6}
		\tilde{\bf p}_i = \sum^{N_c}_{j=1} {\bf{w}}^{p}_{i,j} \otimes \dot{\bp}_{i,j}.
\end{equation}
${\bf{w}}^{{p}}_{i,j}$ is a $3 \times 1$ vector obtained by using the candidate features in a similar way as~\eqref{eq:eq3}, but without the masking operation. By exploiting the expanded candidate points with learnable weights, we obtain more reliable and precise reconstruction results.

\subsection{End-to-End Training}
The proposed network is trained in an end-to-end manner by employing a learnable module for point downsampling.~However, we use two entropy encoders for the sampled position $\bp^*$ and its aggregated feature ${\bf f}^*$, respectively, and the conventional entropy encoders include non-differentiable quantization process which causes the challenge to achieve fully end-to-end training.~Recent advances in learning-based image compression techniques~\cite{end,hyperprior} have paved the way for integrating the entropy encoders into the learning pipeline, and we also adopted the technique~\cite{hyperprior} to approximate the non-differentiable quantization process.  

To train the network, we use the rate loss and the distortion loss as shown in Figure~\ref{fig:fig2}. The rate loss $R$ is defined as the sum of the rates spent to encode the quantized point $\hat{\bp}^*$ and the quantized feature $\hat{\bf f}^*$ given by
\begin{equation}
	R = {\rm E}_{{\bp}}[-\log_{2}q_{{\bp}}(\hat{\bp}^*)] + {\rm E}_{{\bf f}}[-\log_{2}q_{{\bf f}}(\hat{\bf f}^*)],
\end{equation}
where $q_{{\bp}}(\cdot)$ and $q_{{\bf f}}(\cdot)$ denote the probability density functions.~We define the distortion loss $D$ as the Chamfer Distance (CD) loss~\cite{muscle} between the reconstructed point clouds and the original point clouds.~Then the total loss function $L$ is given by 
\begin{equation} \label{eq:eq7}
	L = D + \lambda R,
\end{equation}
where $\lambda$ is the coefficient to determine the contribution of rate loss.

\begin{figure}[t]
  \centering
  \subfloat[]{\includegraphics[width=0.5\linewidth]{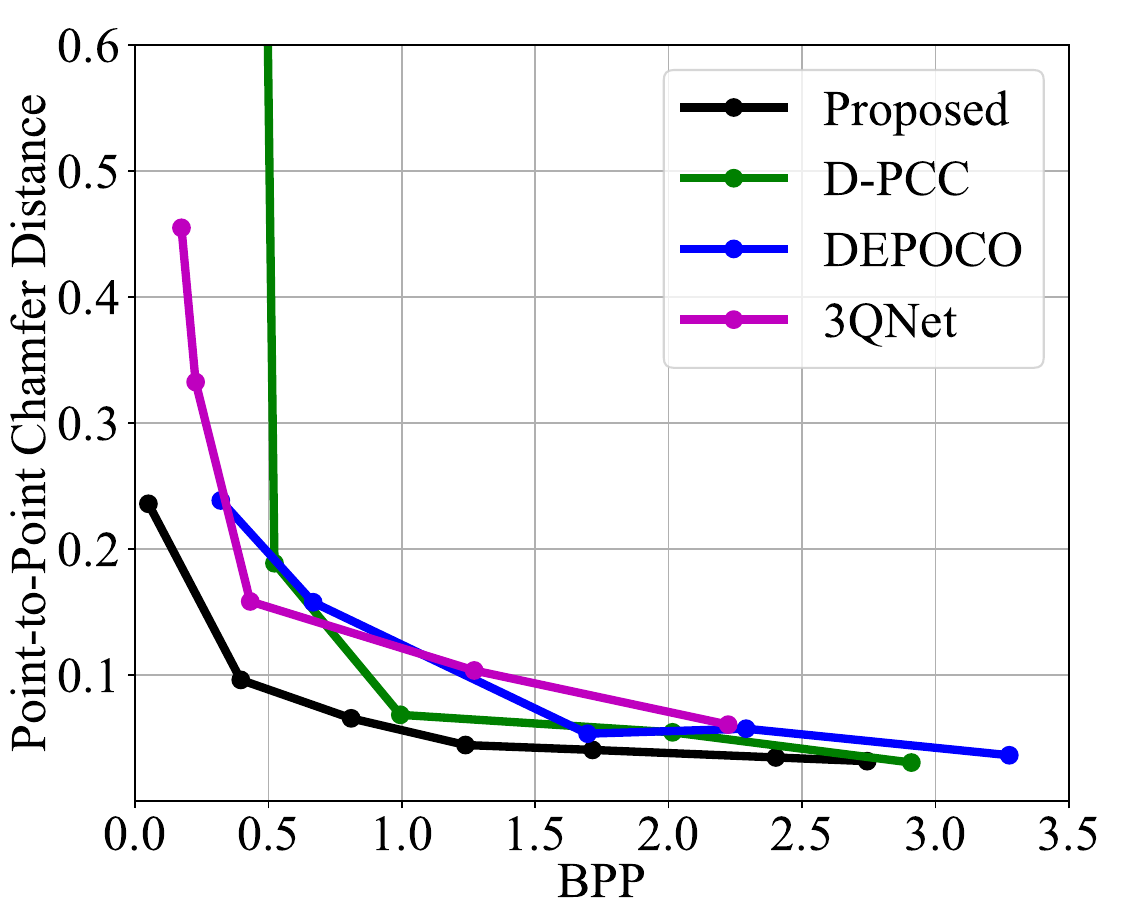}}
  \subfloat[]{\includegraphics[width=0.5\linewidth]{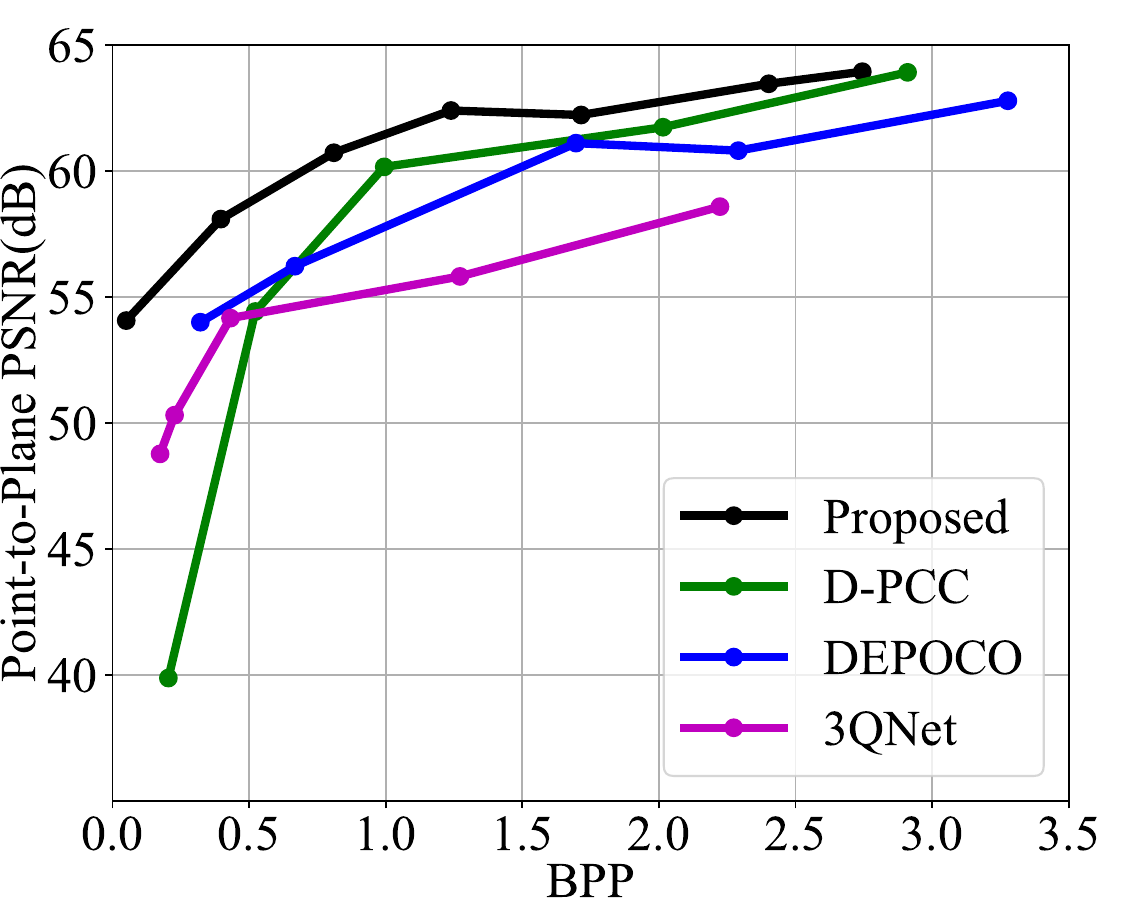}} \vspace{4mm}  \\
  \subfloat[]{\includegraphics[width=0.5\linewidth]{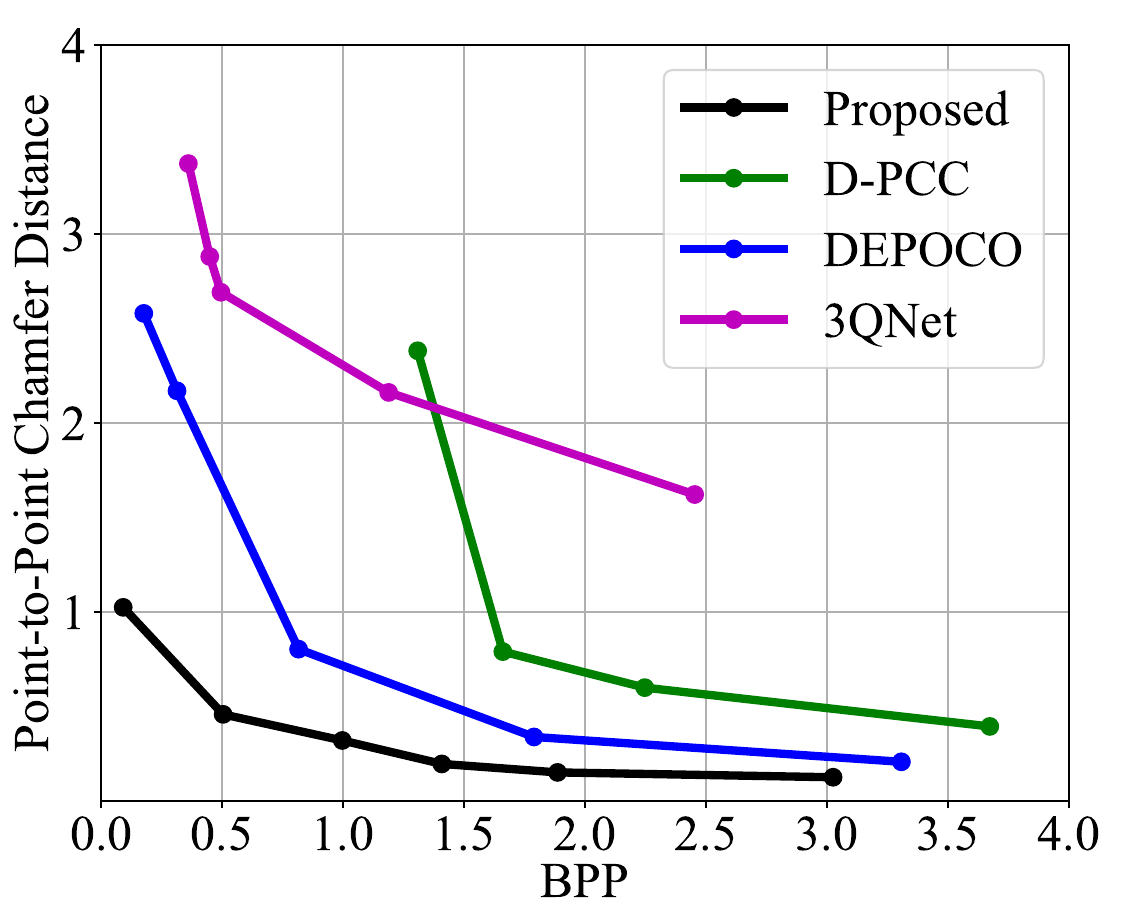}}
  \subfloat[]{\includegraphics[width=0.5\linewidth]{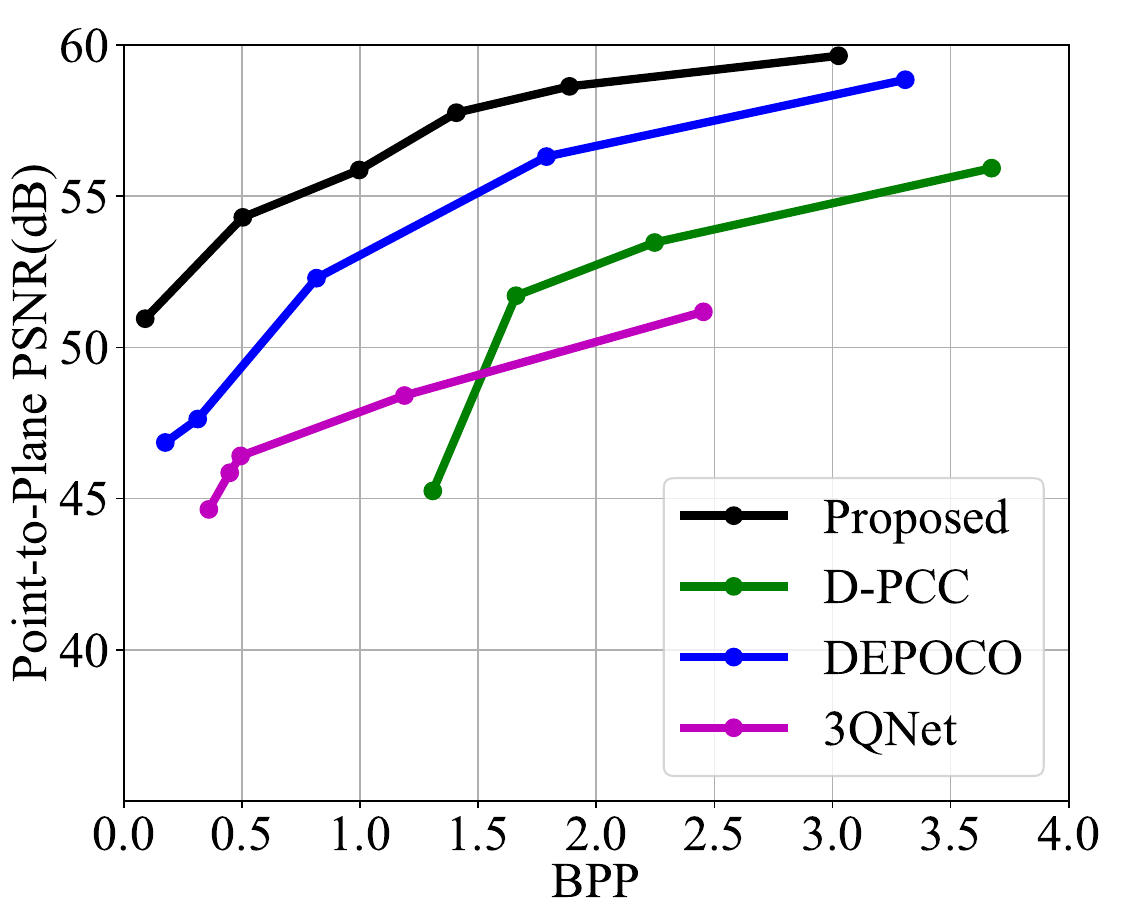}}
  \caption{Rate and distortion curves of the proposed method compared with that of the three state-of-the-art methods: D-PCC~\cite{dpcc}, 3QNet~\cite{3qnet}, and DEPOCO~\cite{depoco}. (a) Bitrate vs.~CD curves and (b) Bitrate vs.~PSNR curves are evaluated on SemanticKITTI dataset. (c) Bitrate vs.~CD curves and (d) Bitrate vs.~PSNR curves are evaluated on nuScenes dataset. The training and test data are from the same dataset.}
  \label{fig:fig5}
\end{figure}

It is worth to note that the point downsampling module in the existing methods is not trainable, and therefore the rate and distortion losses are back-propagated up to the feature extraction module. It means that the compression performance depends on the extracted features only. On the contrary, the proposed method achieves fully end-to-end training by facilitating the loss back-propagation to the learnable point downsampling module. Consequently, the proposed method optimizes not only the extracted features but also the positions of the sampled points simultaneously in terms of the rate and distortion losses.

\section{Experimental Results}



\subsection{Implementation Details}
The performance of the proposed method was evaluated on SemanticKITTI~\cite{semantickitti} and nuScenes~\cite{nuscenes} datasets including quantitative and qualitative comparisons with the existing methods.~SemanticKITTI provides 64-channel LiDAR-based point clouds data of 22 sequences, where we used 10 sequences for training and one sequence for testing as defined in SemanticKITTI.~nuScenes provides 32-channel LiDAR-based point clouds where we used the parts from 1 to 10 with the annotated data only.~Also, we employed the Adam optimizer~\cite{adam} for setting the learning rate to 1e-3.~The range image of the SemanticKITTI dataset has the size of $64 \times 2048$, which was partitioned to local patches of the sizes of $(32 \times 64)$, $(4 \times 16)$, $(4 \times 8)$, $(2 \times 8)$, and $(4 \times 4)$, respectively, to control the bitrate.~The size of the range image of the nuScenes dataset is $32 \times 1072$, and the sizes of local patches were set to $(16 \times 64)$ , $(4 \times 16)$, $(4 \times 8)$, $(2 \times 8)$, and $(2 \times 4)$, respectively.

\begin{figure}[t]
  \centering
  \subfloat[]{\includegraphics[width=0.5\linewidth]{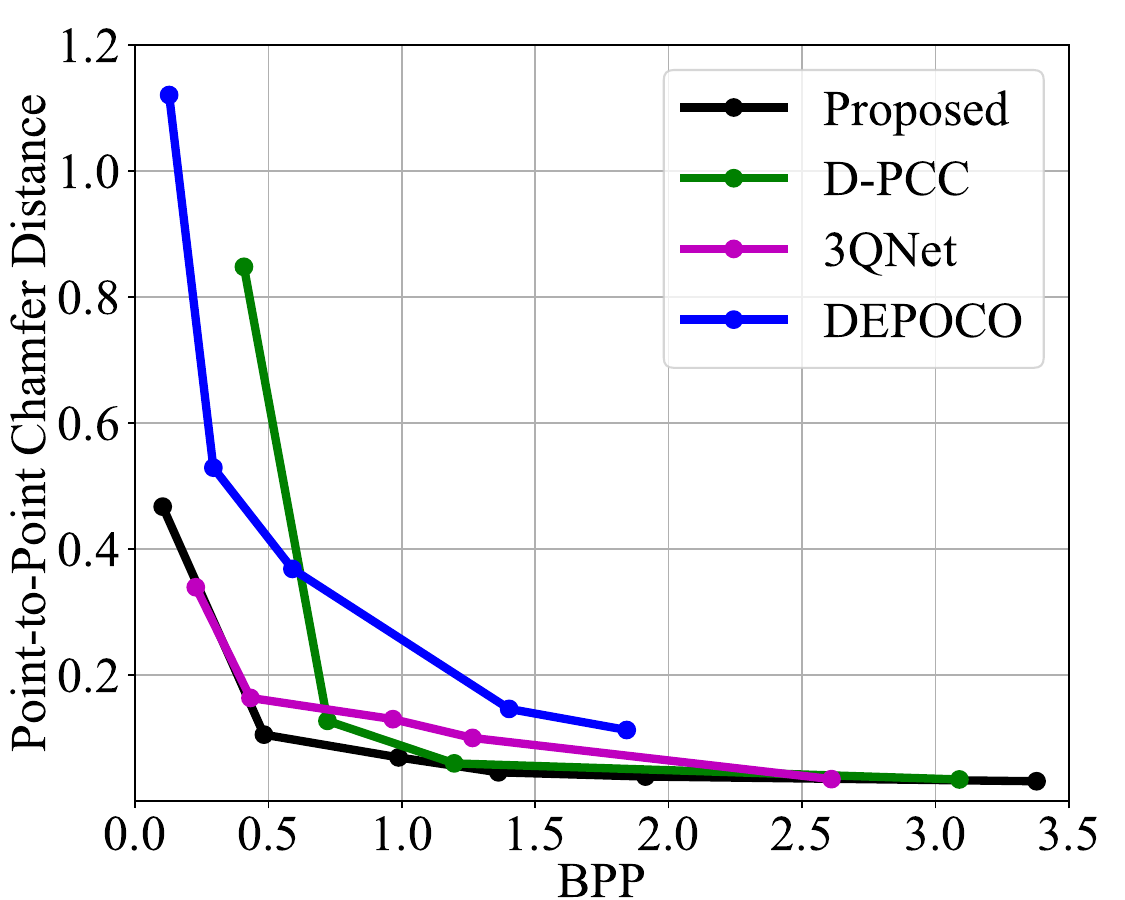}}
  \subfloat[]{\includegraphics[width=0.5\linewidth]{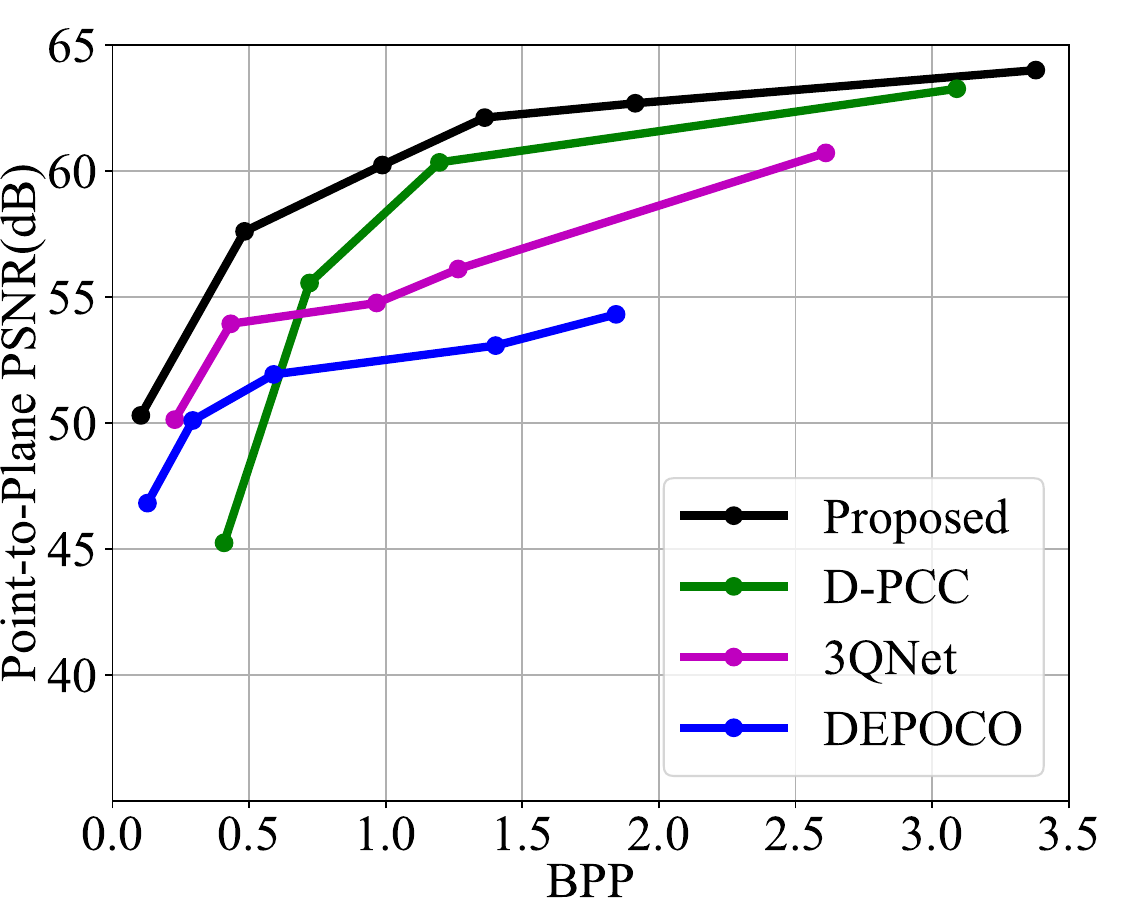}} \vspace{4mm} \\
  \subfloat[]{\includegraphics[width=0.5\linewidth]{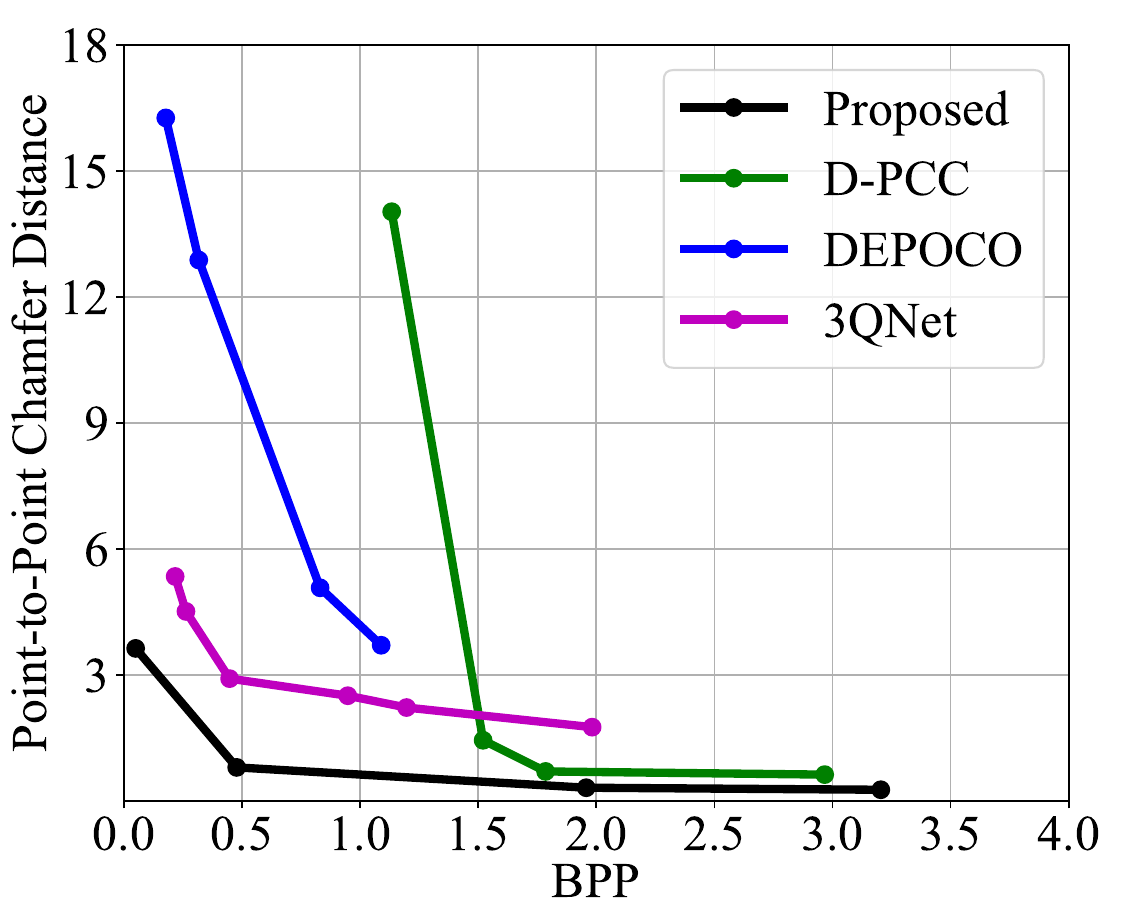}}
  \subfloat[]{\includegraphics[width=0.5\linewidth]{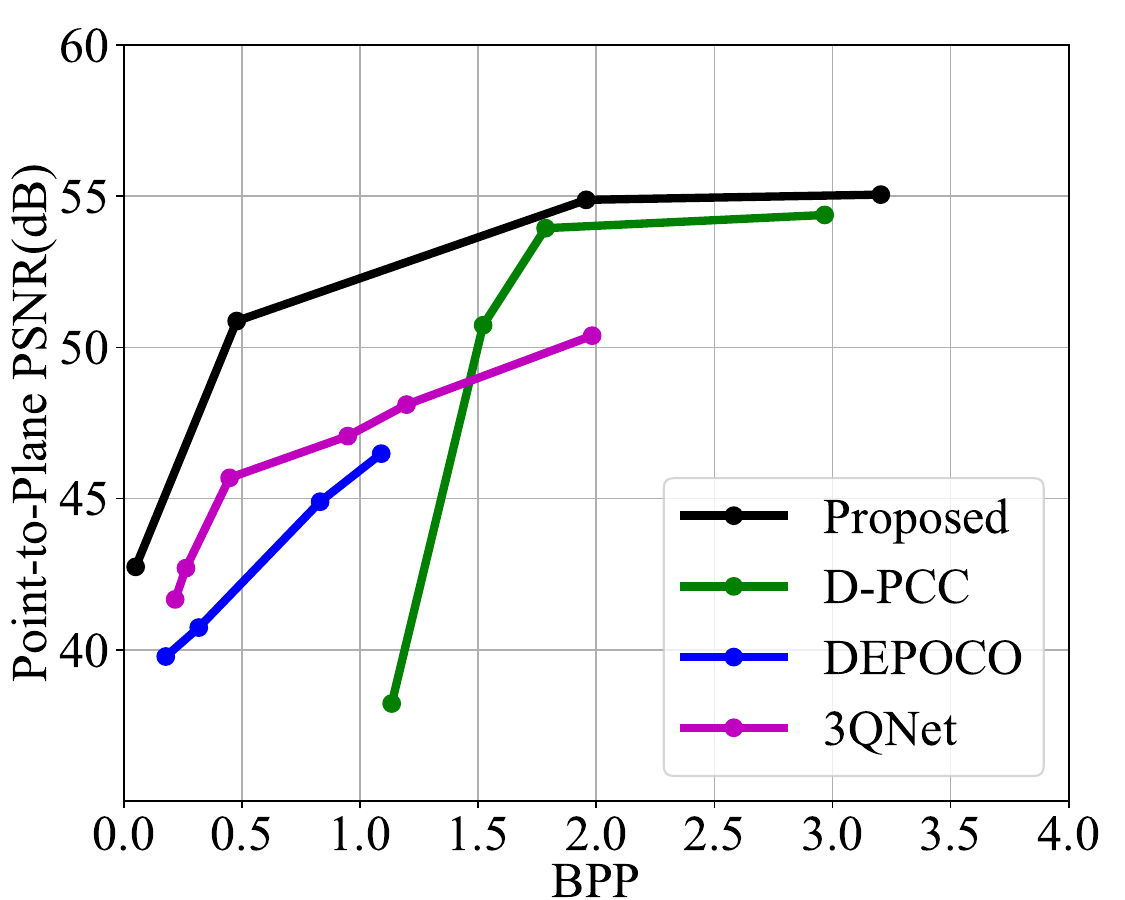}}
  \caption{Rate and distortion curves of the proposed method compared with that of the three state-of-the-art methods: D-PCC~\cite{dpcc}, 3QNet~\cite{3qnet}, and DEPOCO~\cite{depoco}. (a) Bitrate vs.~CD curves and (b) Bitrate vs.~PSNR curves are evaluated on SemanticKITTI dataset. (c) Bitrate vs.~CD curves and (d) Bitrate vs.~PSNR curves are evaluated on nuScenes dataset. The training and test data are from different datasets from each other.}
  \label{fig:fig6}
\end{figure}


\begin{figure*}[t]
  \centering
  \subfloat[Ground Truth]{\includegraphics{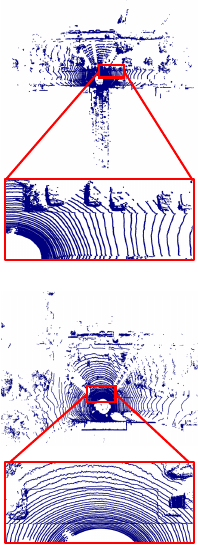}}
  \subfloat[Proposed]{\includegraphics{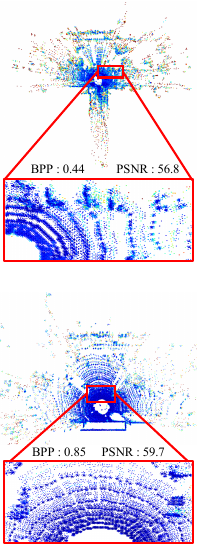}}
  \subfloat[D-PCC~\cite{dpcc}]{\includegraphics{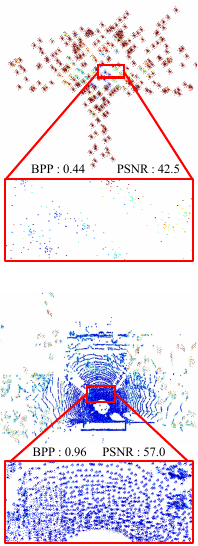}}
  \subfloat[3QNet~\cite{3qnet}]{\includegraphics{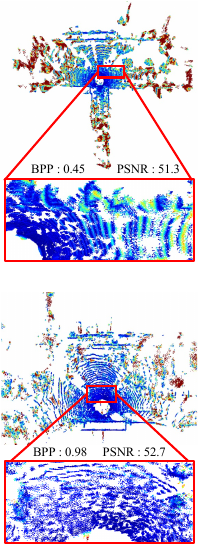}}
  \subfloat[DEPOCO~\cite{depoco}]{\includegraphics{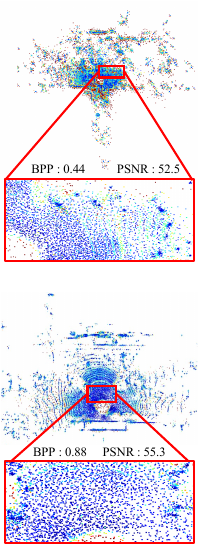}} \\
  \subfloat{\includegraphics{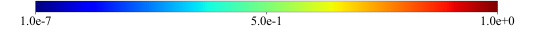}}
  \caption{Comparison of the reconstructed 3D point clouds in SemanticKITTI~\cite{semantickitti} dataset, which are encoded by using the proposed method and three existing methods at similar bitrates. (a) The ground truth point clouds, and the reconstructed point clouds by using (b) the proposed method, (c) D-PCC~\cite{dpcc}, (d) 3QNet~\cite{3qnet}, and (e) DEPOCO~\cite{depoco}. Each point in the reconstructed point clouds is colored according to the distance to the nearest point in the ground truth point clouds.}
  \label{fig:fig7}
\end{figure*}


\subsection{Quantitative Comparison}
\label{sec:quan}
We compared the performance of the proposed method with that of three state-of-the-art methods: D-PCC~\cite{dpcc}, 3QNet~\cite{3qnet}, and DEPOCO~\cite{depoco}. 
Following~\cite{muscle, dpcc}, we measure the reconstruction errors by using two metrics: point-to-point Chamfer distance (CD)~\cite{octsqueeze} and point-to-plane PSNR~\cite{psnr}.


\subsubsection{In-Dataset Evaluation} 
Figure~\ref{fig:fig5} compares the quantitative compression performance in terms of the PSNR and CD curves according to the bits per point (BPP), when the training and test data belong to the same dataset: SemanticKITTI (Figures~\ref{fig:fig5} (a) and (b)) and nuScenes (Figures~\ref{fig:fig5} (c) and (d)) datasets. 
D-PCC~\cite{dpcc} divides the point cloud model into multiple blocks as preprocessing where the outlier blocks, containing smaller numbers of points than a pre-defined threshold, are removed from training. Due to ineffective handling of such outliers, D-PCC has relatively worse performance at low bitrates especially on the nuScenes dataset which has sparser point clouds compared to the SemanticKITTI dataset. 3QNet~\cite{3qnet} randomly upsample the points to ensure the same number of points in each block, and thus generates redundant points that degrade the overall performance on both datasets. DEPOCO~\cite{depoco} yields better performance compared with D-PCC and 3QNet. However it primarily learns the global structure by encoding the entire point cloud model at once without partitioning, causing limited performance of detail reconstruction.
On the other hand, the proposed method not only addresses the aforementioned issues of the existing methods, but leverages a fully end-to-end framework to extract optimal deep features in terms of the rate and distortion. Therefore, we see that the proposed method dramatically outperforms all the compared methods on both of the SemanticKITTI and nuScenes datasets at various bitrates. It demonstrates that the proposed method is a promising solution for LS3DPC data compression in diverse environment.


\begin{figure*}[t]
  \centering
  \subfloat[Ground Truth]{\includegraphics{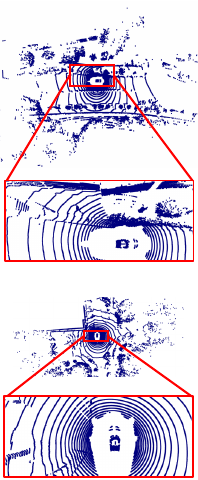}}
  \subfloat[Proposed]{\includegraphics{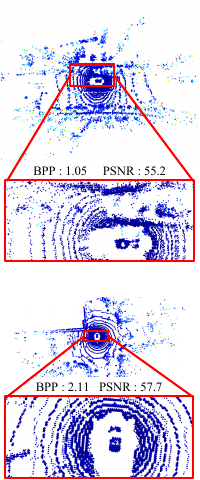}}
  \subfloat[D-PCC~\cite{dpcc}]{\includegraphics{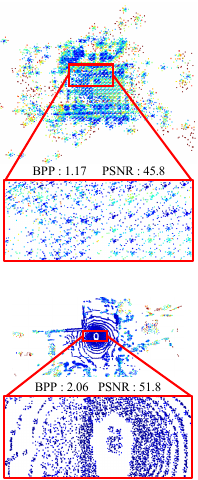}}
  \subfloat[3QNet~\cite{3qnet}]{\includegraphics{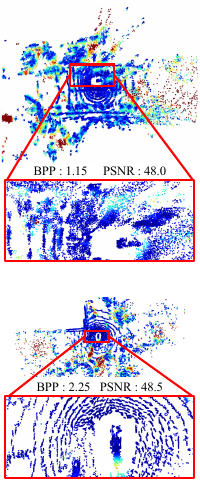}}
  \subfloat[DEPOCO~\cite{depoco}]{\includegraphics{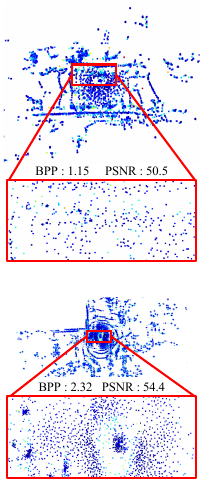}} \\
  \subfloat{\includegraphics{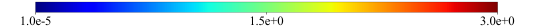}}
  \caption{Comparison of the reconstructed 3D point clouds in nuScenes~\cite{nuscenes} dataset, which are encoded by using the proposed method and three existing methods at similar bitrates. (a) The ground truth point clouds, and the reconstructed point clouds by using (b) the proposed method, (c) D-PCC~\cite{dpcc}, (d) 3QNet~\cite{3qnet}, and (e) DEPOCO~\cite{depoco}. Each point in the reconstructed point clouds is colored according to the distance to the nearest point in the ground truth point clouds.}
  \label{fig:fig8}
\end{figure*}

\subsubsection{Cross-Dataset Evaluation}
Figure~\ref{fig:fig6} also evaluates the generalization capability of the proposed method compared with the existing methods where the training and test data belong to different datasets from each other.~Specifically, the methods were first trained on the nuScenes dataset and then tested on the SemanticKITTI dataset (Figures (a) and (b)), and vice versa (Figures (c) and (d)). In such cases, it is crucial for the network to extract generalized geometric features during training. From this perspective, the results of D-PCC and 3QNet show consistent trends between in-dataset and cross-dataset evaluations, indicating that the partitioning supports robust generalization. However, note that DEPOCO~\cite{depoco} performs the training on the entire point cloud model without partitioning that worsens the generalization ability, and therefore yields significant performance drop with the cross-dataset evaluation though it achieves favorable results with the in-dataset evaluation.
In contrary, the proposed method also achieves the best performance in the cross-dataset evaluation framework demonstrating better generalization capability compared with the existing methods. 



\subsection{Qualitative Comparison}
\label{sec:qual}
We compare the quality of the reconstructed 3D point clouds, which are encoded by using the proposed method and three existing methods at similar bitrates, based on in-dataset evaluation.~Figure~\ref{fig:fig7} shows the results on the SemanticKITTI~\cite{semantickitti} dataset when BPP are below 1. As shown in the first row, the 3D point clouds reconstructed by D-PCC~\cite{dpcc} are barely recognizable, yielding catastrophic distortion at a low bitrate. With a higher bitrate in the second row, the reconstruction quality improves, but we still observe inaccurate regions especially near the center of scanner. 3QNet~\cite{3qnet} shows severe artifacts in the reconstructed point clouds due to the inaccurately predicted redundant points. DEPOCO~\cite{depoco} suffers from the loss of local details, since the network learns the overall shape of the point cloud model as a whole without partitioning. Compared with that of the existing methods, the proposed method provides more faithful results and shows significantly better quality of the reconstructed point clouds especially at low bitrates. 

\begin{table}[t]
	\caption{Comparison of the computation complexity in terms of the memory space to store the parameters of network and the time of encoding and decoding executed on SemanticKITTI dataset by using NVIDIA TITAN RTX GPUs.}
	\centering
	\begin{tabular}{ccccc}
		\toprule
		\multirow{2}{*}{\raisebox{-0.45\height}{Methods}} & \multirow{2}{*}{\raisebox{-0.65\height}{\begin{tabular}[c]{@{}c@{}}Memory\\ (MB) \end{tabular}}} & \multicolumn{3}{c}{Time (ms)} \\ \cmidrule(lr){3-5} 
		&                & \multicolumn{1}{c}{Enc.}     & \multicolumn{1}{c}{Dec.}   & Total  \\ \midrule
		\multicolumn{1}{c|}{D-PCC~\cite{dpcc}}    & {\bf 0.33} & \multicolumn{1}{|c}{1801}    & \multicolumn{1}{c}{817}   & 2618     \\ 
		\multicolumn{1}{c|}{3QNet~\cite{3qnet}}   & 27.08      & \multicolumn{1}{|c}{584}     & \multicolumn{1}{c}{862}   & 1446     \\ 
		\multicolumn{1}{c|}{DEPOCO~\cite{depoco}} & 0.47       & \multicolumn{1}{|c}{345}     & \multicolumn{1}{c}{\bf 3} & 348      \\ 
		\multicolumn{1}{c|}{Proposed}             & 3.36       & \multicolumn{1}{|c}{\bf 56}  & \multicolumn{1}{c}{12}    & {\bf 68} \\ \bottomrule
	\end{tabular}
	\label{tab:tab1}
\end{table}

Figure~\ref{fig:fig8} also compares the reconstructed point clouds in the nuScenes~\cite{nuscenes} dataset, where the bitrates are above 1 BPP. As shown in the second row, D-PCC exhibits relatively huge distortions at outlier regions far from the center of scanner due to ineffective handling of outliers. 3QNet yields the distortion near and far regions together, though it preserves some local details. Compared to other methods, DEPOCO accurately reconstructs the outlier regions and exhibits relatively smaller distortions. However, it loses local details in the reconstructed point clouds. On the contrary, the proposed method accurately restores the original point clouds while preserving the intricate geometric details, outperforming all the compared methods in terms of the reconstruction quality at similar bitrates.

\subsection{Computational Complexity} 
We also investigated the computational complexity of the proposed method. Table~\ref{tab:tab1} compares the memory space to store the model's parameters, as well as the average execution time required for encoding and decoding, based on the experiments conducted on the SemanticKITTI dataset. D-PCC has the smallest model size, but spends the largest time since it divides the entire point cloud model into smaller blocks, which are processed sequentially. On the other hand, 3QNet requires the largest memory space about 27MB to store the model parameters, and also incurs a high time cost. DEPOCO processes the entire scene data at once with downsampling, and yields a small network size. Also, it achieves the fastest decoding time. However, note that all the compared existing methods suffer from large encoding times mainly due to time-consuming operations including FPS and KNN. Though a relatively large model size, the proposed method achieves the smallest encoding time as well as the smallest overall time, since it partitions the point cloud model on the range image domain, enabling it to process the entire scene at once.

\subsection{Ablation Study}
\label{sec:ablation}
\subsubsection{Effect of Learnable Point Sampling}

\begin{figure}[t]
	\centering
	\subfloat[]{\includegraphics[width=0.5\linewidth]{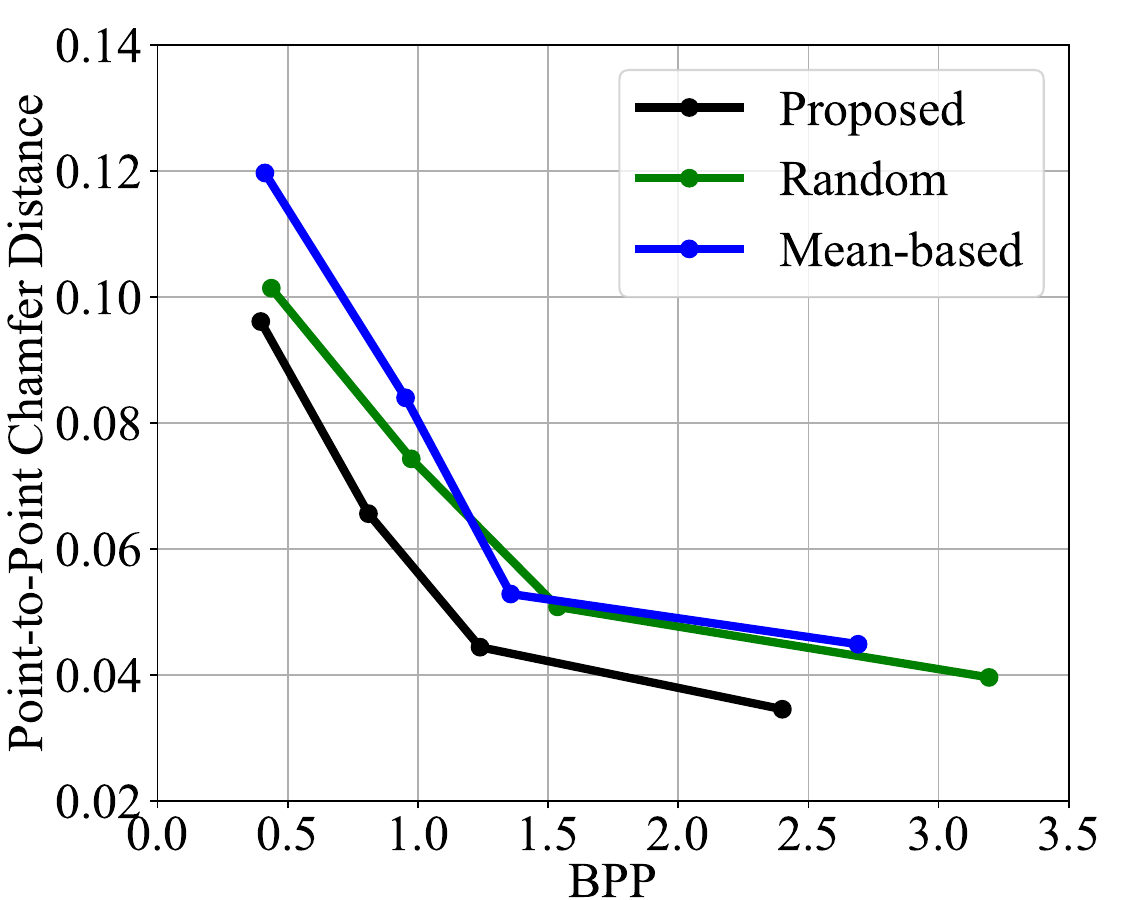}}
	\subfloat[]{\includegraphics[width=0.5\linewidth]{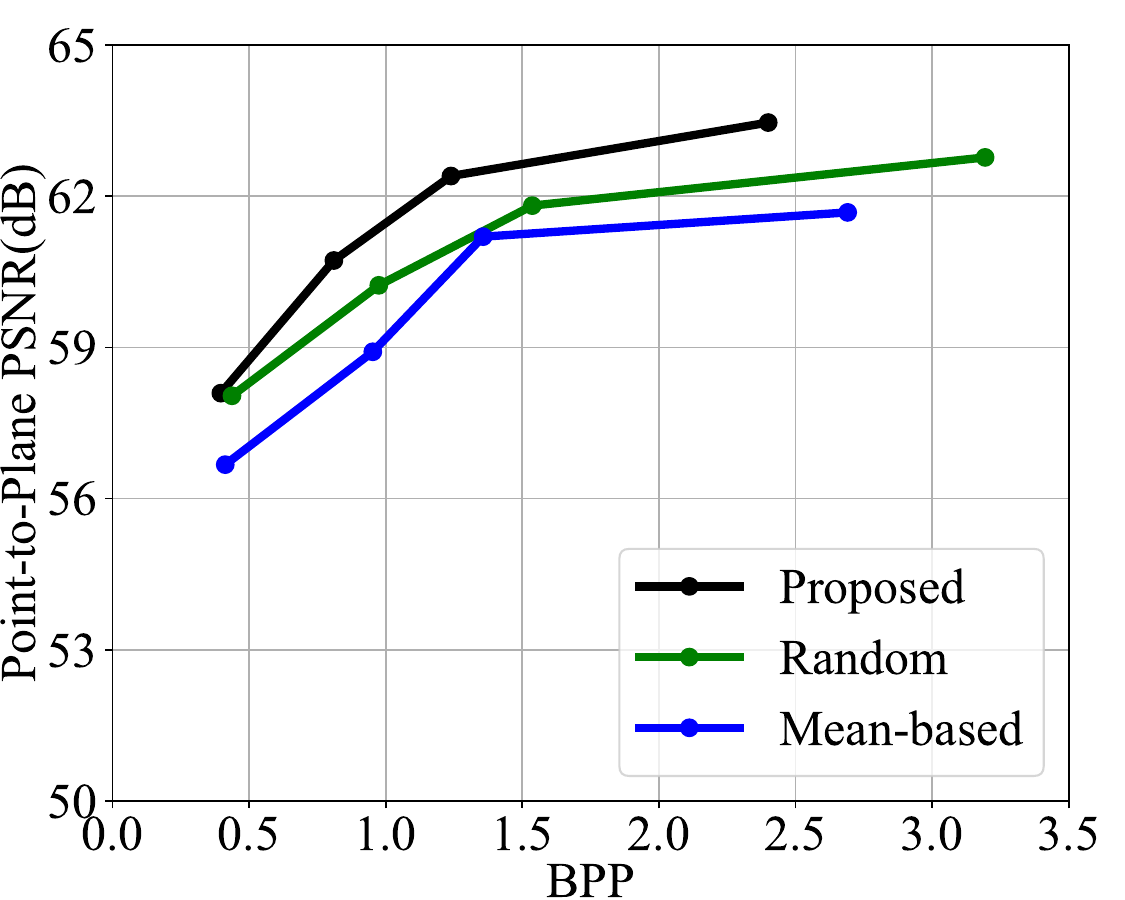}}
	\caption{Effect of the proposed learnable weights-based point sampling compared to the random sampling and the mean-based sampling.~(a) Bitrate vs.~CD curves and (b) Bitrate vs.~PSNR curves, evaluated on SemanticKITTI dataset.}
	\label{fig:fig9}
\end{figure}

Figure~\ref{fig:fig9} shows the effect of the proposed point downsampling with learnable weights.~The random sampling randomly selects one of the original points as a downsampled point, and the mean-based sampling takes the average location of the points within each patch.~We see that the fully end-to-end training framework using the learnable point sampling method shows better quality of the reconstructed models, while spending less bits for encoding. This demonstrates that the learnable point sampling is important to achieve better compression ratios, as it determines the optimal location of downsampled point by minimizing the rate and distortion losses.

\subsubsection{Effect of Decoder Architecture}

\begin{figure}[t]
	\centering
	\subfloat[]{\includegraphics[width=0.5\linewidth]{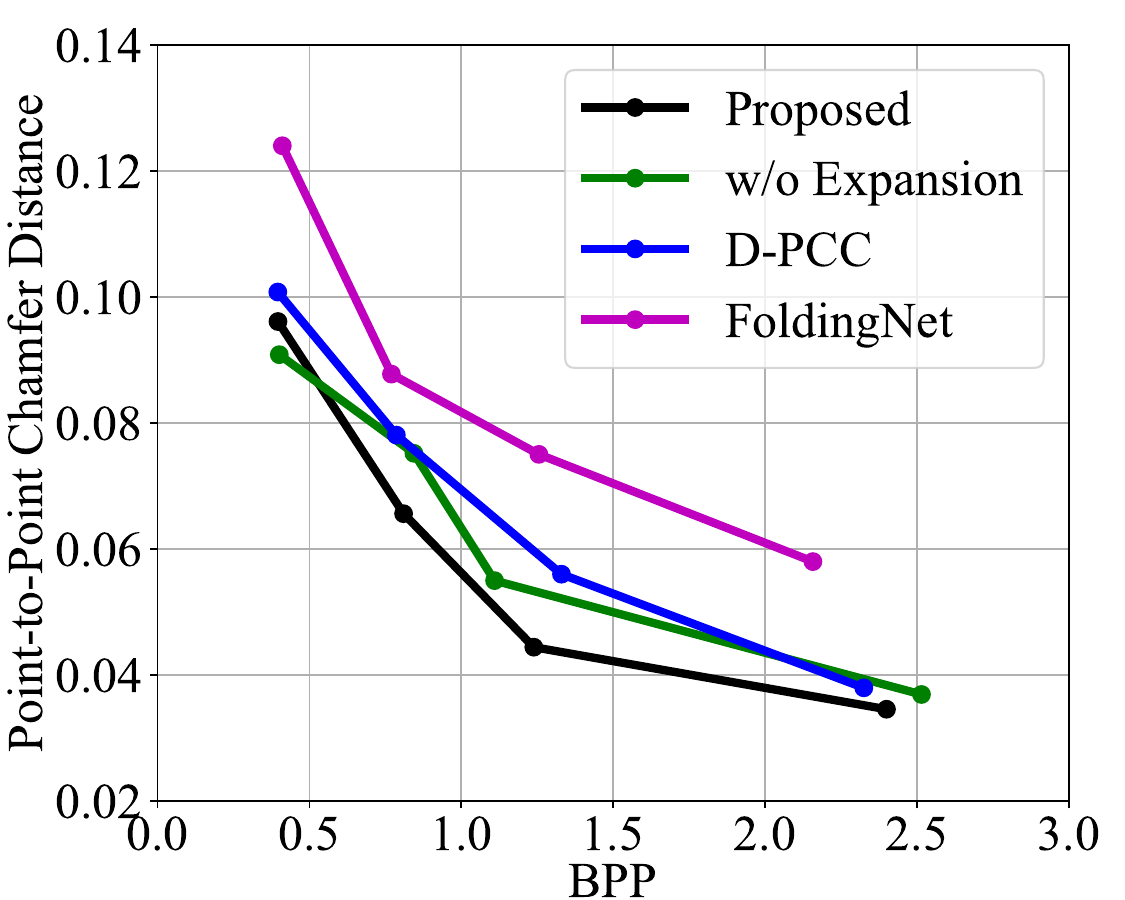}}
	\subfloat[]{\includegraphics[width=0.5\linewidth]{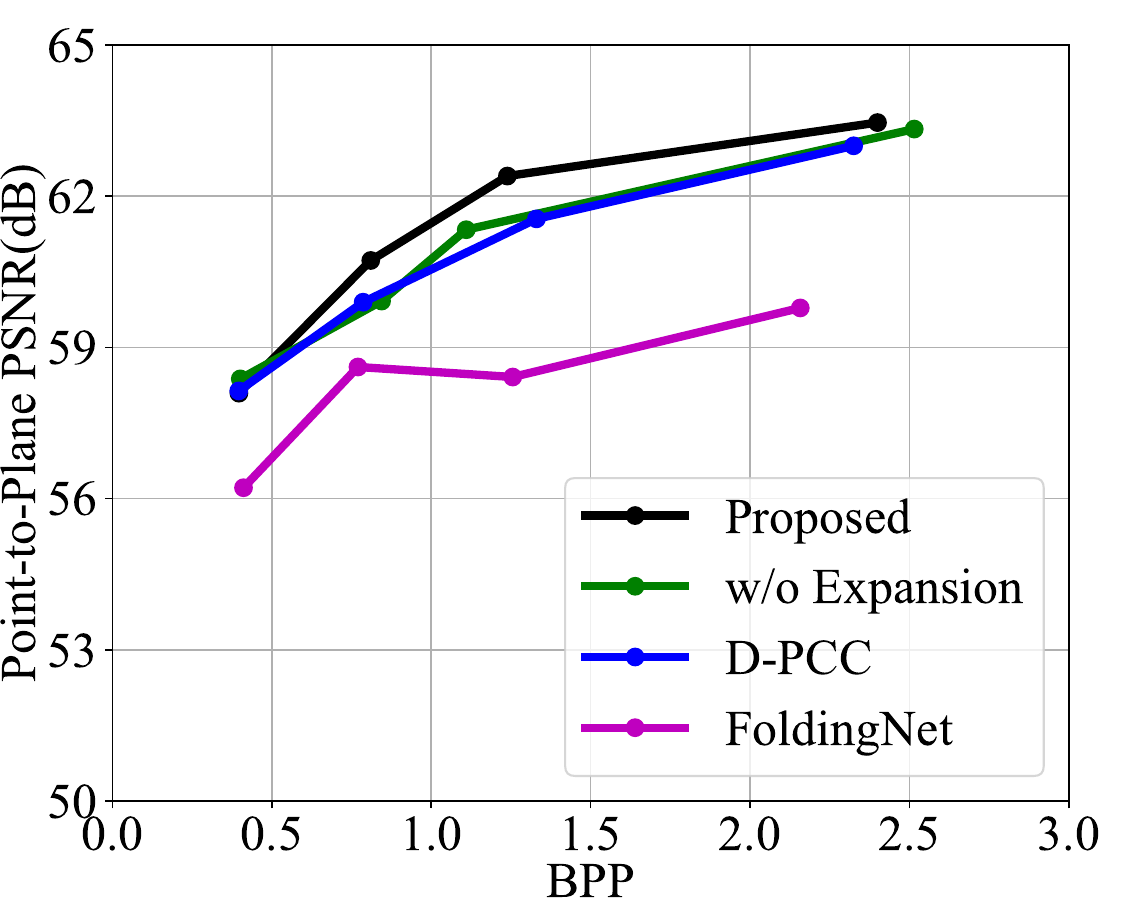}}
	\caption{Effect of the proposed expansion and fusion based decoder compared to the existing methods. (a) Bitrate vs.~CD curves and (b) Bitrate vs.~PSNR curves, evaluated on SemanticKITTI dataset.}
	\label{fig:fig10}
\end{figure}

Figure~\ref{fig:fig10} also validates the effect of the expansion and fusion module in the proposed decoder architecture. 'w/o Expansion' means the proposed decoder with the expansion ratio of 1 (${N}_c=1$). Moreover, for comparison to the existing methods, we replaced the proposed decoder with that of two established methods: FoldingNet~\cite{foldingnet} and D-PCC~\cite{dpcc}, respectively. We observe that the proposed decoder achieves comparable performance to D-PCC even without expansion. With the expansion, fusing multiple candidate points to refine the reconstructed points greatly improves the performance.

\section{Conclusion}
In this paper, we proposed a novel compression framework for LS3DPC where the point sampling and feature extraction are jointly optimized. An input LS3DPC model is partitioned on the range image domain, and each partition is downsampled into a single point, respectively. We devised a learnable point sampling network which is jointly optimized with the feature extraction in an end-to-end manner, thereby estimating an optimal location of the downsampled point as well as aggregating an optimal feature in terms of the rate and distortion losses. Moreover, we designed an expansion and fusion module for point refinement network that adaptively aggregates expanded candidate points to reconstruct the original points reliably. Experimental results demonstrated that the proposed method achieves superior rate-distortion performance compared with the existing state-of-the-art methods. 

{
    \small
    \bibliographystyle{ieeenat_fullname}
    \bibliography{main}

\begin{thebibliography}{49}
\providecommand{\natexlab}[1]{#1}
\providecommand{\url}[1]{\texttt{#1}}
\expandafter\ifx\csname urlstyle\endcsname\relax
  \providecommand{\doi}[1]{doi: #1}\else
  \providecommand{\doi}{doi: \begingroup \urlstyle{rm}\Url}\fi

\bibitem[Ahn et~al.(2015)Ahn, Lee, Sim, and Kim]{rangetra1}
Jae-Kyun Ahn, Kyu-Yul Lee, Jae-Young Sim, and Chang-Su Kim.
\newblock Large-scale 3d point cloud compression using adaptive radial distance
  prediction in hybrid coordinate domains.
\newblock \emph{IEEE J. Sel. Topics Signal Process.}, 9\penalty0 (3):\penalty0
  422--434, 2015.

\bibitem[Ball{\'e} et~al.(2017)Ball{\'e}, Laparra, and Simoncelli]{end}
Johannes Ball{\'e}, Valero Laparra, and Eero~P Simoncelli.
\newblock End-to-end optimized image compression.
\newblock In \emph{ICLR}, pages 1--27, 2017.

\bibitem[Ballé et~al.(2018)Ballé, Minnen, Singh, Hwang, and
  Johnston]{hyperprior}
Johannes Ballé, David Minnen, Saurabh Singh, Sung~Jin Hwang, and Nick
  Johnston.
\newblock Variational image compression with a scale hyperprior.
\newblock In \emph{ICLR}, pages 1--23, 2018.

\bibitem[Behley et~al.(2019)Behley, Garbade, Milioto, Quenzel, Behnke,
  Stachniss, and Gall]{semantickitti}
Jens Behley, Martin Garbade, Andres Milioto, Jan Quenzel, Sven Behnke, Cyrill
  Stachniss, and Jurgen Gall.
\newblock Semantickitti: A dataset for semantic scene understanding of lidar
  sequences.
\newblock In \emph{ICCV}, pages 9297--9307, 2019.

\bibitem[Biswas et~al.(2020)Biswas, Liu, Wong, Wang, and Urtasun]{muscle}
Sourav Biswas, Jerry Liu, Kelvin Wong, Shenlong Wang, and Raquel Urtasun.
\newblock Muscle: Multi sweep compression of lidar using deep entropy models.
\newblock In \emph{NeurIPS}, pages 22170--22181, 2020.

\bibitem[Bretar and Chehata(2009)]{terrain}
Fr{\'e}d{\'e}ric Bretar and Nesrine Chehata.
\newblock Terrain modeling from lidar range data in natural landscapes: A
  predictive and bayesian framework.
\newblock \emph{IEEE Trans. Geosci. Remote Sens.}, 48\penalty0 (3):\penalty0
  1568--1578, 2009.

\bibitem[Caesar et~al.(2020)Caesar, Bankiti, Lang, Vora, Liong, Xu, Krishnan,
  Pan, Baldan, and Beijbom]{nuscenes}
Holger Caesar, Varun Bankiti, Alex~H Lang, Sourabh Vora, Venice~Erin Liong,
  Qiang Xu, Anush Krishnan, Yu Pan, Giancarlo Baldan, and Oscar Beijbom.
\newblock nuscenes: A multimodal dataset for autonomous driving.
\newblock In \emph{CVPR}, pages 11621--11631, 2020.

\bibitem[Cui et~al.(2023)Cui, Long, Feng, Li, and Kai]{octformer}
Mingyue Cui, Junhua Long, Mingjian Feng, Boyang Li, and Huang Kai.
\newblock Octformer: Efficient octree-based transformer for point cloud
  compression with local enhancement.
\newblock In \emph{AAAI}, pages 470--478, 2023.

\bibitem[Dosovitskiy et~al.(2021)Dosovitskiy, Beyer, Kolesnikov, Weissenborn,
  Zhai, Unterthiner, Dehghani, Minderer, Heigold, Gelly, et~al.]{vit}
Alexey Dosovitskiy, Lucas Beyer, Alexander Kolesnikov, Dirk Weissenborn,
  Xiaohua Zhai, Thomas Unterthiner, Mostafa Dehghani, Matthias Minderer, Georg
  Heigold, Sylvain Gelly, et~al.
\newblock An image is worth 16x16 words: Transformers for image recognition at
  scale.
\newblock In \emph{ICLR}, pages 1--11, 2021.

\bibitem[Fan et~al.(2023)Fan, Gao, Xu, Wang, and Li]{multiscale}
Tingyu Fan, Linyao Gao, Yiling Xu, Dong Wang, and Zhu Li.
\newblock Multiscale latent-guided entropy model for lidar point cloud
  compression.
\newblock \emph{IEEE TCSVT}, 33\penalty0 (12):\penalty0 7857--7869, 2023.

\bibitem[Galligan et~al.(2018)Galligan, Hemmer, Stava, Zhang, and
  Brettle]{draco}
Frank Galligan, Michael Hemmer, Ondrej Stava, Fan Zhang, and Jamieson Brettle.
\newblock Google/draco: a library for compressing and decompressing 3d
  geometric meshes and point clouds, 2018.

\bibitem[Graziosi et~al.(2020)Graziosi, Nakagami, Kuma, Zaghetto, Suzuki, and
  Tabatabai]{gpcc}
Danillo Graziosi, Ohji Nakagami, Satoru Kuma, Alexandre Zaghetto, Teruhiko
  Suzuki, and Ali Tabatabai.
\newblock An overview of ongoing point cloud compression standardization
  activities: Video-based (v-pcc) and geometry-based (g-pcc).
\newblock In \emph{APSIPA Trans. Signal Inf. Process.}, page e13, 2020.

\bibitem[He et~al.(2022)He, Ren, Tang, Zhang, Xue, and Fu]{dpcc}
Yun He, Xinlin Ren, Danhang Tang, Yinda Zhang, Xiangyang Xue, and Yanwei Fu.
\newblock Density-preserving deep point cloud compression.
\newblock In \emph{CVPR}, pages 2333--2342, 2022.

\bibitem[Houshiar and N{\"u}chter(2015)]{rangetra2}
Hamidreza Houshiar and Andreas N{\"u}chter.
\newblock 3d point cloud compression using conventional image compression for
  efficient data transmission.
\newblock In \emph{XXV Int. Conf. Inf., Commun. Autom. Technol. (ICAT)}, pages
  1--8, 2015.

\bibitem[Huang et~al.(2020)Huang, Wang, Wong, Liu, and Urtasun]{octsqueeze}
Lila Huang, Shenlong Wang, Kelvin Wong, Jerry Liu, and Raquel Urtasun.
\newblock Octsqueeze: Octree-structured entropy model for lidar compression.
\newblock In \emph{CVPR}, pages 1313--1323, 2020.

\bibitem[Huang and Liu(2019)]{pointnet2based}
Tianxin Huang and Yong Liu.
\newblock 3d point cloud geometry compression on deep learning.
\newblock pages 890--898, 2019.

\bibitem[Huang et~al.(2022)Huang, Zhang, Chen, Ding, Tai, Zhang, Wang, and
  Liu]{3qnet}
Tianxin Huang, Jiangning Zhang, Jun Chen, Zhonggan Ding, Ying Tai, Zhenyu
  Zhang, Chengjie Wang, and Yong Liu.
\newblock 3qnet: 3d point cloud geometry quantization compression network.
\newblock \emph{ACM TOG}, 41\penalty0 (6):\penalty0 1--13, 2022.

\bibitem[Kingma and Ba(2014)]{adam}
Diederik~P Kingma and Jimmy Ba.
\newblock Adam: A method for stochastic optimization.
\newblock \emph{arXiv preprint arXiv:1412.6980}, 2014.

\bibitem[Lang et~al.(2019)Lang, Vora, Caesar, Zhou, Yang, and
  Beijbom]{pointpillars}
Alex~H Lang, Sourabh Vora, Holger Caesar, Lubing Zhou, Jiong Yang, and Oscar
  Beijbom.
\newblock Pointpillars: Fast encoders for object detection from point clouds.
\newblock In \emph{CVPR}, pages 12697--12705, 2019.

\bibitem[Luo et~al.(2024)Luo, Song, Nonaka, Unno, Sun, Goto, and Katto]{scp}
Ao Luo, Linxin Song, Keisuke Nonaka, Kyohei Unno, Heming Sun, Masayuki Goto,
  and Jiro Katto.
\newblock Scp: Spherical-coordinate-based learned point cloud compression.
\newblock In \emph{AAAI}, pages 3954--3962, 2024.

\bibitem[Mekuria et~al.(2016)Mekuria, Blom, and Cesar]{traditional1}
Rufael Mekuria, Kees Blom, and Pablo Cesar.
\newblock Design, implementation, and evaluation of a point cloud codec for
  tele-immersive video.
\newblock \emph{IEEE TCSVT}, 27\penalty0 (4):\penalty0 828--842, 2016.

\bibitem[Minnen et~al.(2018)Minnen, Ball{\'e}, and Toderici]{joint}
David Minnen, Johannes Ball{\'e}, and George~D Toderici.
\newblock Joint autoregressive and hierarchical priors for learned image
  compression.
\newblock In \emph{NeurIPS}, pages 10794--10803, 2018.

\bibitem[Peixoto(2020)]{dyadic}
Eduardo Peixoto.
\newblock Intra-frame compression of point cloud geometry using dyadic
  decomposition.
\newblock \emph{IEEE Signal Process. Lett.}, 27:\penalty0 246--250, 2020.

\bibitem[Qi et~al.(2017{\natexlab{a}})Qi, Su, Mo, and Guibas]{pointnet}
Charles~R Qi, Hao Su, Kaichun Mo, and Leonidas~J Guibas.
\newblock Pointnet: Deep learning on point sets for 3d classification and
  segmentation.
\newblock In \emph{CVPR}, pages 652--660, 2017{\natexlab{a}}.

\bibitem[Qi et~al.(2017{\natexlab{b}})Qi, Yi, Su, and Guibas]{pointnet2}
Charles~Ruizhongtai Qi, Li Yi, Hao Su, and Leonidas~J Guibas.
\newblock Pointnet++: Deep hierarchical feature learning on point sets in a
  metric space.
\newblock In \emph{NeurIPS}, 2017{\natexlab{b}}.

\bibitem[Quach et~al.(2019)Quach, Valenzise, and Dufaux]{pccgeocnnv1}
Maurice Quach, Giuseppe Valenzise, and Frederic Dufaux.
\newblock Learning convolutional transforms for lossy point cloud geometry
  compression.
\newblock In \emph{ICIP}, pages 4320--4324, 2019.

\bibitem[Quach et~al.(2020)Quach, Valenzise, and Dufaux]{pccgeocnnv2}
Maurice Quach, Giuseppe Valenzise, and Frederic Dufaux.
\newblock Improved deep point cloud geometry compression.
\newblock pages 1--6, 2020.

\bibitem[Que et~al.(2021)Que, Lu, and Xu]{voxelcontext}
Zizheng Que, Guo Lu, and Dong Xu.
\newblock Voxelcontext-net: An octree based framework for point cloud
  compression.
\newblock In \emph{CVPR}, pages 6042--6051, 2021.

\bibitem[Schnabel and Klein(2006)]{traditional2}
Ruwen Schnabel and Reinhard Klein.
\newblock Octree-based point-cloud compression.
\newblock In \emph{Proc. 3rd Symp. Point-Based Graph.}, pages 111--120, 2006.

\bibitem[Song et~al.(2021)Song, Shao, Gao, Wang, and Li]{layerwise}
Fei Song, Yiting Shao, Wei Gao, Haiqiang Wang, and Thomas Li.
\newblock Layer-wise geometry aggregation framework for lossless lidar point
  cloud compression.
\newblock \emph{IEEE TCSVT}, 31\penalty0 (12):\penalty0 4603--4616, 2021.

\bibitem[Song et~al.(2023)Song, Fu, Liu, and Li]{efficient}
Rui Song, Chunyang Fu, Shan Liu, and Ge Li.
\newblock Efficient hierarchical entropy model for learned point cloud
  compression.
\newblock In \emph{CVPR}, pages 14368--14377, 2023.

\bibitem[Thomas et~al.(2019)Thomas, Qi, Deschaud, Marcotegui, Goulette, and
  Guibas]{kpconv}
Hugues Thomas, Charles~R Qi, Jean-Emmanuel Deschaud, Beatriz Marcotegui,
  Fran{\c{c}}ois Goulette, and Leonidas~J Guibas.
\newblock Kpconv: Flexible and deformable convolution for point clouds.
\newblock In \emph{ICCV}, pages 6411--6420, 2019.

\bibitem[Tian et~al.(2017)Tian, Ochimizu, Feng, Cohen, and Vetro]{psnr}
Dong Tian, Hideaki Ochimizu, Chen Feng, Robert Cohen, and Anthony Vetro.
\newblock Geometric distortion metrics for point cloud compression.
\newblock In \emph{ICIP}, pages 3460--3464, 2017.

\bibitem[Vaswani et~al.(2017)Vaswani, Shazeer, Parmar, Uszkoreit, Jones, Gomez,
  Kaiser, and Polosukhin]{transformer}
Ashish Vaswani, Noam Shazeer, Niki Parmar, Jakob Uszkoreit, Llion Jones,
  Aidan~N Gomez, {\L}ukasz Kaiser, and Illia Polosukhin.
\newblock Attention is all you need.
\newblock In \emph{NeurIPS}, pages 5998--6008, 2017.

\bibitem[Wang et~al.(2021{\natexlab{a}})Wang, Zhu, Xu, Xu, and Yang]{pointvote}
Chaofei Wang, Wenjie Zhu, Yingzhan Xu, Yiling Xu, and Le Yang.
\newblock Point-voting based point cloud geometry compression.
\newblock pages 1--5, 2021{\natexlab{a}}.

\bibitem[Wang et~al.(2021{\natexlab{b}})Wang, Ding, Li, and Ma]{pcgcv2}
Jianqiang Wang, Dandan Ding, Zhu Li, and Zhan Ma.
\newblock Multiscale point cloud geometry compression.
\newblock pages 73--82, 2021{\natexlab{b}}.

\bibitem[Wang et~al.(2021{\natexlab{c}})Wang, Zhu, Liu, and Ma]{pcgcv1}
Jianqiang Wang, Hao Zhu, Haojie Liu, and Zhan Ma.
\newblock Lossy point cloud geometry compression via end-to-end learning.
\newblock \emph{IEEE TCSVT}, 31\penalty0 (12):\penalty0 4909--4923,
  2021{\natexlab{c}}.

\bibitem[Wang et~al.(2022)Wang, Ding, Li, Feng, Cao, and Ma]{sparsepcgc}
Jianqiang Wang, Dandan Ding, Zhu Li, Xiaoxing Feng, Chuntong Cao, and Zhan Ma.
\newblock Sparse tensor-based multiscale representation for point cloud
  geometry compression.
\newblock \emph{IEEE TPAMI}, 45\penalty0 (7):\penalty0 9055--9071, 2022.

\bibitem[Wang et~al.(2024)Wang, Huang, Dong, Lin, Song, and Xie]{mslpcc}
Miaohui Wang, Runnan Huang, Hengjin Dong, Di Lin, Yun Song, and Wuyuan Xie.
\newblock mslpcc: A multimodal-driven scalable framework for deep lidar point
  cloud compression.
\newblock In \emph{AAAI}, pages 5526--5534, 2024.

\bibitem[Wang and Liu(2022)]{rangeentropy}
Sukai Wang and Ming Liu.
\newblock Point cloud compression with range image-based entropy model for
  autonomous driving.
\newblock In \emph{ECCV}, pages 323--340, 2022.

\bibitem[Wang et~al.(2019)Wang, Sun, Liu, Sarma, Bronstein, and Solomon]{dgcnn}
Yue Wang, Yongbin Sun, Ziwei Liu, Sanjay~E Sarma, Michael~M Bronstein, and
  Justin~M Solomon.
\newblock Dynamic graph cnn for learning on point clouds.
\newblock \emph{ACM TOG}, 38\penalty0 (5):\penalty0 1--12, 2019.

\bibitem[Wei et~al.(2022)Wei, Niu, Xiao, and He]{isolated}
Ziwei Wei, Benben Niu, Haodong Xiao, and Yun He.
\newblock Isolated points prediction via deep neural network on point cloud
  lossless geometry compression.
\newblock \emph{IEEE TCSVT}, 33\penalty0 (1):\penalty0 407--420, 2022.

\bibitem[Wiesmann et~al.(2021)Wiesmann, Milioto, Chen, Stachniss, and
  Behley]{depoco}
Louis Wiesmann, Andres Milioto, Xieyuanli Chen, Cyrill Stachniss, and Jens
  Behley.
\newblock Deep compression for dense point cloud maps.
\newblock 6\penalty0 (2):\penalty0 2060--2067, 2021.

\bibitem[Yan et~al.(2019)Yan, Liu, Li, Li, Li, et~al.]{pointnetbased}
Wei Yan, Shan Liu, Thomas~H Li, Zhu Li, Ge Li, et~al.
\newblock Deep autoencoder-based lossy geometry compression for point clouds.
\newblock \emph{arXiv preprint arXiv:1905.03691}, 2019.

\bibitem[Yang et~al.(2018)Yang, Feng, Shen, and Tian]{foldingnet}
Yaoqing Yang, Chen Feng, Yiru Shen, and Dong Tian.
\newblock Foldingnet: Point cloud auto-encoder via deep grid deformation.
\newblock In \emph{CVPR}, pages 206--215, 2018.

\bibitem[Zhao et~al.(2021)Zhao, Jiang, Jia, Torr, and Koltun]{pointtransformer}
Hengshuang Zhao, Li Jiang, Jiaya Jia, Philip~HS Torr, and Vladlen Koltun.
\newblock Point transformer.
\newblock In \emph{ICCV}, pages 16259--16268, 2021.

\bibitem[Zhou et~al.(2022)Zhou, Qi, Zhou, and Anguelov]{riddle}
Xuanyu Zhou, Charles~R Qi, Yin Zhou, and Dragomir Anguelov.
\newblock Riddle: Lidar data compression with range image deep delta encoding.
\newblock In \emph{CVPR}, pages 17212--17221, 2022.

\bibitem[Zhu et~al.(2021)Zhu, Xu, Ding, Ma, and Nilsson]{regionpcgc}
Wenjie Zhu, Yiling Xu, Dandan Ding, Zhan Ma, and Mike Nilsson.
\newblock Lossy point cloud geometry compression via region-wise processing.
\newblock \emph{IEEE TCSVT}, 31\penalty0 (12):\penalty0 4575--4589, 2021.

\bibitem[Zou et~al.(2021)Zou, Sun, Chen, Nie, and Li]{navigation}
Qin Zou, Qin Sun, Long Chen, Bu Nie, and Qingquan Li.
\newblock A comparative analysis of lidar slam-based indoor navigation for
  autonomous vehicles.
\newblock \emph{IEEE Trans. Intell. Transp. Syst.}, 23\penalty0 (7):\penalty0
  6907--6921, 2021.

\end{thebibliography}
}


\end{document}